\documentclass[11pt]{article}
\usepackage[margin=1in]{geometry}
\usepackage{amsmath,amssymb,amsthm}
\usepackage{mathtools}
\usepackage{booktabs}
\usepackage{enumitem}
\usepackage[numbers,sort&compress]{natbib}
\usepackage{graphicx}
\usepackage{hyperref}
\usepackage{xcolor}

\graphicspath{{./}}

\definecolor{teal}{HTML}{0D9488}

\newtheorem{definition}{Definition}
\newtheorem{remark}{Remark}

\title{\textbf{Progressive Autonomy as Preference Learning}\\[4pt]
\large A Formalization of Trust Calibration for Agentic Tool Use}
\author{Changkun Ou}
\date{March 4, 2026}

\begin{document}
\maketitle

\begin{abstract}
We formalize trust calibration for agentic tool use (deciding when an
automated agent's proposed action may execute autonomously versus require
human approval) as a preference-learning problem. A policy gateway maintains
a Gaussian-process posterior over a latent human risk-tolerance function,
observed through a probit likelihood on binary approve/deny feedback, and
escalates to the human exactly where the approval outcome is most uncertain.
We show this is structurally an instance of Preferential Bayesian
Optimization, inheriting its inference machinery (approximate Gaussian-process
classification) and its sample-efficiency argument (uncertainty-targeted
querying), while differing in objective: classifying an action space into
allow/block/ask regions rather than optimizing a design.
\end{abstract}

\section{Background and Related Work}

Deciding how much autonomy to delegate to an automated system is a classical
human-in-the-loop control problem; the technical core here, recovering a
human's latent acceptability function from sparse binary feedback, is
\emph{preference learning}. \citet{chu2005preference} introduced
Gaussian-process preference learning, placing a Gaussian-process (GP) prior
over a latent utility and linking observed human choices to it through a
probit likelihood. We adopt exactly this construction, specialized to unary
approve/deny feedback. The same latent-utility-with-probit model underlies
preference-driven sequential decision making: \citet{gonzalez2017preferential}
formalized \emph{Preferential Bayesian Optimization} (PBO), embedding GP
preference learning in the query loop of Bayesian optimization
\citep{brochu2010tutorial,shahriari2016bayesopt}. Our policy gateway is
structurally an instance of this framework, differing only in objective
(classifying an action space rather than optimizing a design), as
Table~\ref{tab:mapping} makes precise.

The inferential machinery is classical GP classification: a GP prior, a
non-Gaussian probit likelihood, and an analytically intractable posterior
approximated by the Laplace method or Expectation Propagation
\citep{minka2001ep}, developed comprehensively by
\citet{rasmussen2006gaussian}. Treating the human query budget as a scarce
resource makes the \textsc{ask} region an \emph{acquisition} rule in the sense
of active learning \citep{settles2009active} and Bayesian optimization
\citep{shahriari2016bayesopt}: interruptions are spent where the expected
information about the allow/block boundary is greatest. Finally, drifting risk
tolerance is a non-stationarity problem; we model it with a time-decaying
kernel component in the spirit of non-stationary covariance functions
\citep{paciorek2003nonstationary} and, for abrupt shifts, Bayesian online
changepoint detection \citep{adams2007bocd}.

This progressive view of autonomy has deep roots in the trust-in-automation
literature: \citet{lee2004trust} characterized appropriate reliance as the
alignment of trust with actual system trustworthiness, and
\citet{devisser2020longitudinal} extended this to \emph{longitudinal} trust
calibration in human--robot teams, precisely the dynamic our time-decaying
kernel (\S\ref{sec:nonstationary}) is meant to capture. The concern has
resurfaced sharply for large-language-model agents, where graduated autonomy
has emerged as an explicit deployment axis alongside raw capability
\citep{morris2024levels}. Recent work catalogs the risks of increasingly
agentic systems \citep{chan2023harms}, argues for visibility and oversight
mechanisms over deployed agents \citep{chan2024visibility}, and proposes
governance practices in which a human retains approval authority over
consequential actions \citep{shavit2023practices}. These accounts are largely
qualitative and taxonomic: they argue \emph{why} graduated autonomy matters
and \emph{what} should be governed, but leave the escalation policy itself as
a fixed, hand-specified tier. Our contribution is the missing mechanism: a
learning rule that lets the auto-approve/escalate boundary adapt from human
feedback rather than being set by hand.

\section{Setup}

At each decision point $t = 1, 2, \ldots$, the policy gateway observes a proposed agent action $a_t \in \mathcal{A}$ and an execution context $c_t \in \mathcal{C}$, where:
\begin{align}
    a_t &= (\texttt{tool\_name},\; \texttt{args},\; \texttt{target\_resource}), \\
    c_t &= (\texttt{repo\_state},\; \texttt{task\_desc},\; \texttt{session\_history}).
\end{align}
A human supervisor provides binary feedback $y_t \in \{0, 1\}$ (deny/approve). We write $x_t \coloneqq (a_t, c_t) \in \mathcal{X} = \mathcal{A} \times \mathcal{C}$ for the joint input.

\section{Latent Risk Tolerance}

\begin{definition}[Risk tolerance function]
There exists a latent function $f\colon \mathcal{X} \to \mathbb{R}$ encoding the human's risk tolerance, such that approval probability follows a probit observation model:
\begin{equation}
    \Pr(y = 1 \mid x) = \Phi\!\bigl(f(x)\bigr),
\end{equation}
where $\Phi$ is the standard normal CDF.
\end{definition}

\begin{remark}\label{rem:unary}
This is structurally identical to the observation model in Preferential Bayesian Optimization \citep{gonzalez2017preferential} and, more fundamentally, in Gaussian-process preference learning \citep{chu2005preference}. In standard PBO, the human expresses a preference between two candidates $(x_i, x_j)$ via $\Pr(x_i \succ x_j) = \Phi\bigl(f(x_i) - f(x_j)\bigr)$. Here, the comparison is against an implicit internal threshold: the human approves when $f(x)$ exceeds their acceptable-risk boundary. The unary case is a degenerate pairwise comparison against a fixed reference point.
\end{remark}

\section{Gaussian Process Prior and Posterior}

Place a GP prior over $f$ \citep{rasmussen2006gaussian}:
\begin{equation}
    f \sim \mathcal{GP}\!\bigl(\mu_0,\; k(x, x')\bigr).
\end{equation}

The kernel $k$ decomposes over the input structure. A natural choice is a product kernel:
\begin{equation}
    k(x, x') = k_{\text{tool}}(a, a') \cdot k_{\text{ctx}}(c, c') \cdot k_{\text{time}}(t, t'),
\end{equation}
where:
\begin{itemize}[nosep]
    \item $k_{\text{tool}}$ encodes similarity between actions (e.g., shared tool name, overlapping argument patterns, same reversibility class),
    \item $k_{\text{ctx}}$ captures context similarity (same repository, file type, task category),
    \item $k_{\text{time}}$ handles non-stationarity (see \S\ref{sec:nonstationary}).
\end{itemize}

After observing $\mathcal{D}_N = \{(x_t, y_t)\}_{t=1}^{N}$, the posterior is:
\begin{equation}\label{eq:posterior}
    p(f \mid \mathcal{D}_N) \;\propto\; \mathcal{GP}(\mu_0, k) \;\cdot \prod_{t=1}^{N} \Phi\!\bigl(f(x_t)\bigr)^{y_t} \bigl(1 - \Phi\!\bigl(f(x_t)\bigr)\bigr)^{1 - y_t}.
\end{equation}

This is analytically intractable due to the non-Gaussian likelihood. Approximate inference proceeds via the Laplace approximation \citep{rasmussen2006gaussian} or Expectation Propagation \citep{minka2001ep}, the same machinery used in PBO.

\begin{remark}[Practical realization]
For the unary model, the Laplace approximation of \citet{rasmussen2006gaussian} (Algorithm~3.1 there) reduces inference to a short Newton iteration to the posterior mode, after which the predictive approval probability admits the closed form $\mathbb{E}\bigl[\Phi(f(x_*))\bigr] \approx \Phi\!\bigl(\mu_* / \sqrt{1 + \sigma_*^2}\bigr)$, with $\mu_*$ and $\sigma_*^2$ the latent posterior mean and variance at $x_*$. At the scale of decision points in a single project ($N$ on the order of $10^2$ to $10^3$), this exact computation is inexpensive and needs no sparse approximation, which keeps the implementation a transparent instance of the cited machinery rather than a wrapper around a general optimization stack. A maintained off-the-shelf alternative is the \texttt{PairwiseGP} model of BoTorch \citep{balandat2020botorch}, a deployable PBO implementation; being pairwise, it recovers the unary setting only through the degenerate comparison against a fixed reference of Remark~\ref{rem:unary}, at the cost of a heavier dependency. This choice affects only the posterior: the gateway rule, the non-stationary drift component, and any baselines are unchanged.
\end{remark}

\section{The Policy Gateway Decision Rule}

Given the posterior predictive distribution at a new point $x_*$:
\begin{equation}
    \hat{p}(x_*) \coloneqq \mathbb{E}_{f \mid \mathcal{D}_N}\!\bigl[\Phi(f(x_*))\bigr],
\end{equation}
the gateway applies a three-tier decision:
\begin{equation}\label{eq:decision}
    \text{decision}(x_*) =
    \begin{cases}
        \textsc{allow}  & \text{if } \hat{p}(x_*) > \tau_{\text{high}}, \\[3pt]
        \textsc{block}  & \text{if } \hat{p}(x_*) < \tau_{\text{low}}, \\[3pt]
        \textsc{ask}    & \text{otherwise}.
    \end{cases}
\end{equation}

The $\textsc{ask}$ region $[\tau_{\text{low}}, \tau_{\text{high}}]$ plays the role of an \emph{acquisition function} \citep{shahriari2016bayesopt,settles2009active}: the system queries the human precisely where the model is most uncertain about the approval outcome, maximizing the expected value of information per human interruption.

\begin{remark}
The target operating point (85--90\% auto-approve, 10--15\% human escalation) emerges naturally as the posterior concentrates. Early in a project, most actions fall in the $\textsc{ask}$ region. As the model learns, the $\textsc{ask}$ band narrows, and the auto-approve rate rises without any manual threshold tuning.
\end{remark}

\section{Non-Stationarity}\label{sec:nonstationary}

Human risk tolerance drifts: early in a project the supervisor is cautious; as trust accumulates for familiar patterns, they become permissive. Model this via a time-decayed kernel component:
\begin{equation}
    k_{\text{time}}(t, t') = \exp\!\Bigl(-\frac{|t - t'|}{\lambda}\Bigr),
\end{equation}
where $\lambda > 0$ is a lengthscale controlling the forgetting rate. This gives recent approve/deny signals more weight, a principled analog of the intuition that ``agents earn trust over time.''

For computational efficiency, an equivalent effect can be achieved through a sliding window of the most recent $W$ observations, or via online changepoint detection when the supervisor's behavior shifts abruptly (e.g., moving to a new codebase).

\section{Correlated Generalization}

A key advantage over a na\"ive contextual bandit (which treats each $(a, c)$ independently) is \emph{correlated generalization} through the kernel. Concretely:

\begin{itemize}[nosep]
    \item Approving \texttt{write\_file} to \texttt{/workspace/src/} transfers evidence to \texttt{write\_file} to \texttt{/workspace/test/}, since $k_{\text{tool}}$ and $k_{\text{ctx}}$ assign high similarity.
    \item Denying \texttt{execute\_sql} with a \texttt{DROP} argument propagates caution to \texttt{execute\_sql} with \texttt{TRUNCATE}, without the human having to deny each variant individually.
    \item A new tool with no interaction history inherits the prior $\mu_0$, which maps to $\textsc{ask}$, the fail-safe default.
\end{itemize}

\section{Connection to PBO}

The mapping between the trust calibration problem and Preferential Bayesian Optimization \citep{gonzalez2017preferential} is summarized in Table~\ref{tab:mapping}.

\begin{table}[h]
\centering
\begin{tabular}{@{}lll@{}}
\toprule
\textbf{Component} & \textbf{PBO (Optimization)} & \textbf{Trust Calibration (Policy)} \\
\midrule
Input space $\mathcal{X}$ & Design parameters & (action, context) pairs \\
Latent function $f$ & Objective to maximize & Risk tolerance to learn \\
Human feedback & Pairwise preference $x_i \succ x_j$ & Unary approve/deny \\
Observation model & $\Phi(f(x_i) - f(x_j))$ & $\Phi(f(x))$ \\
Acquisition & Next query to evaluate & Next action to escalate \\
Goal & Find $x^* = \arg\max f$ & Learn the $\textsc{allow}$/$\textsc{block}$ boundary \\
\bottomrule
\end{tabular}
\caption{Structural correspondence between PBO and trust calibration.}
\label{tab:mapping}
\end{table}

The ``preferential'' aspect is literal: the human is expressing preferences over what the agent should be allowed to do. The same mathematical machinery (GP priors, probit likelihoods, approximate posterior inference) transfers directly. The difference is the objective: PBO seeks to \emph{optimize}, while trust calibration seeks to \emph{classify} the action space into allow/block/ask regions with minimal human queries.

\section{Datasets and Evaluation}

Empirically grounding the gateway requires data pairing proposed agent
actions with human approve/deny judgments. The closest public resource is
R-Judge \citep{yuan-etal-2024-rjudge}, which supplies multi-turn agent
interaction records annotated by humans with binary safe/unsafe labels across
a range of risk scenarios; it is a natural source for a cold-start prior
$\mu_0$ and for calibrating $k_{\text{tool}}$ and $k_{\text{ctx}}$. Broader
agent-safety benchmarks such as Agent-SafetyBench
\citep{zhang2024agentsafetybench} and ToolEmu \citep{ruan2024toolemu} widen
the action and context coverage, but their risk labels are produced by
automated judges rather than per-action human approval, so they are better
suited to stress-testing the learned boundary than to fitting $f$ itself.

A structural gap remains: no public dataset captures the \emph{longitudinal,
per-supervisor} signal that the non-stationarity model
(\S\ref{sec:nonstationary}) assumes. Existing corpora provide single-shot,
aggregated annotations and do not track an individual supervisor's risk
tolerance drifting over the course of a project. Validating the time-decaying
kernel therefore requires either a controlled user study or a simulation with
deliberately drifting synthetic annotators, with R-Judge serving as the
static initializer. We treat this as an inherent limitation of currently
available data rather than a deficiency of the model.

\section{Simulation Study}\label{sec:simulation}

We accordingly exercise the formulation in a controlled simulation whose
ground-truth oracle instantiates Definition~1 and the drift of
\S\ref{sec:nonstationary}: the standard evaluation protocol for Preferential
Bayesian Optimization, where the latent preference function is synthetic by
construction so that recovery, calibration, and query efficiency can be
measured against a known target. The action space is $18$ agent tools with
interpretable decision-time risk attributes (reversibility, base sensitivity,
blast radius, destructive-argument flag), $8$ target-resource sensitivity
tiers, and $7$ task contexts. The oracle latent is a static action
acceptability term plus a saturating accumulated-trust term
(\S\ref{sec:nonstationary}) with an abrupt changepoint at $t = 750$, plus a
three-way safety veto (irreversible \emph{and} sensitive \emph{and}
low-trust); approvals are drawn as $y \sim \mathrm{Bernoulli}(\Phi(f^*))$. The
gateway is the self-contained Laplace GP-probit of \citet{rasmussen2006gaussian}
with the product kernel of \S4. The kernel observes the action and context
risk attributes but never the time-varying veto or the drift, which it must
recover from feedback through $k_{\text{time}}$. We process a stream of
$N = 1500$ decision points over $6$ seeds, prequentially (every decision is
scored before any label at that step exists), with phases learn $[0, 560)$,
validation $[560, 1050)$ where $(\tau_{\text{low}}, \tau_{\text{high}})$ is
tuned once under a tightened false-allow cap, and a frozen-policy test
$[1050, 1500)$ in which the model keeps adapting online. Reproducible code,
the full report, and all figures are provided under \texttt{experiment/}.

\begin{figure}[tb]
\centering
\includegraphics[width=0.49\linewidth]{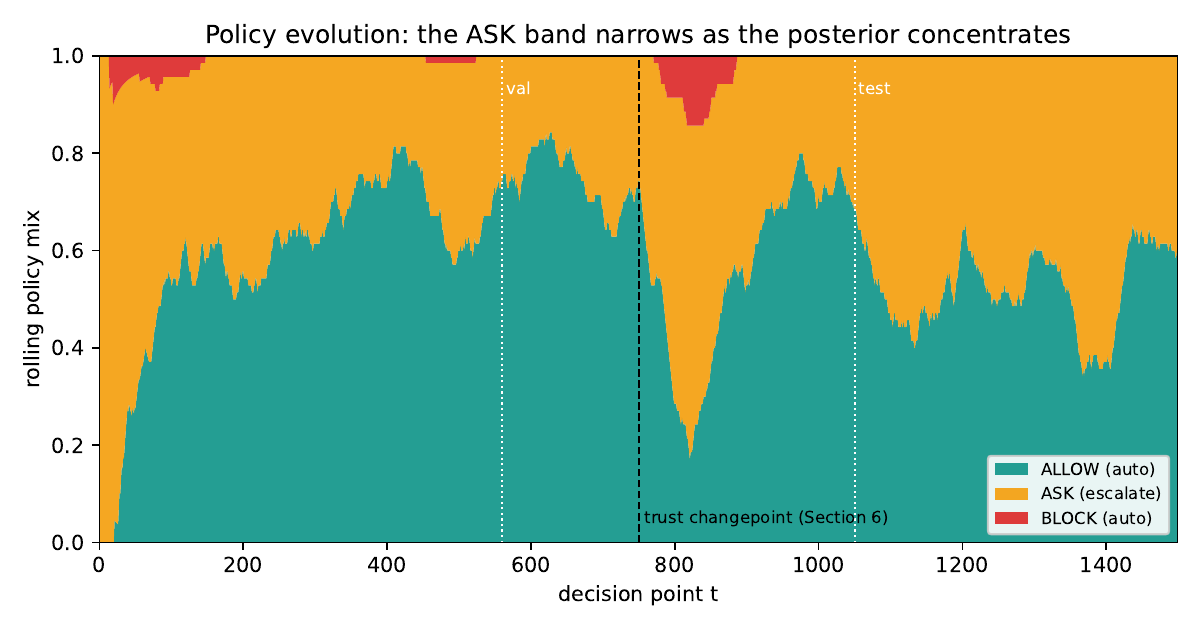}\hfill
\includegraphics[width=0.49\linewidth]{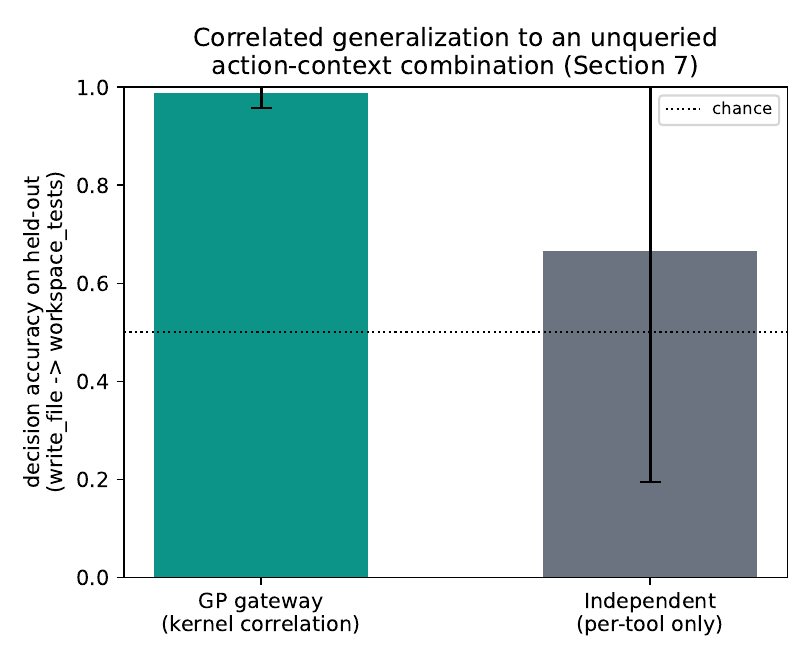}
\caption{Left: the rolling policy mix. The \textsc{allow} share rises as the
posterior concentrates (the \textsc{ask} band narrows), then collapses at the
$t=750$ trust changepoint and recovers, exactly the Section~5 and Section~6
dynamics. Right: correlated generalization (Section~7) to an action-context
combination that was never queried.}
\label{fig:gateway}
\end{figure}

\begin{figure}[tb]
\centering
\includegraphics[width=0.49\linewidth]{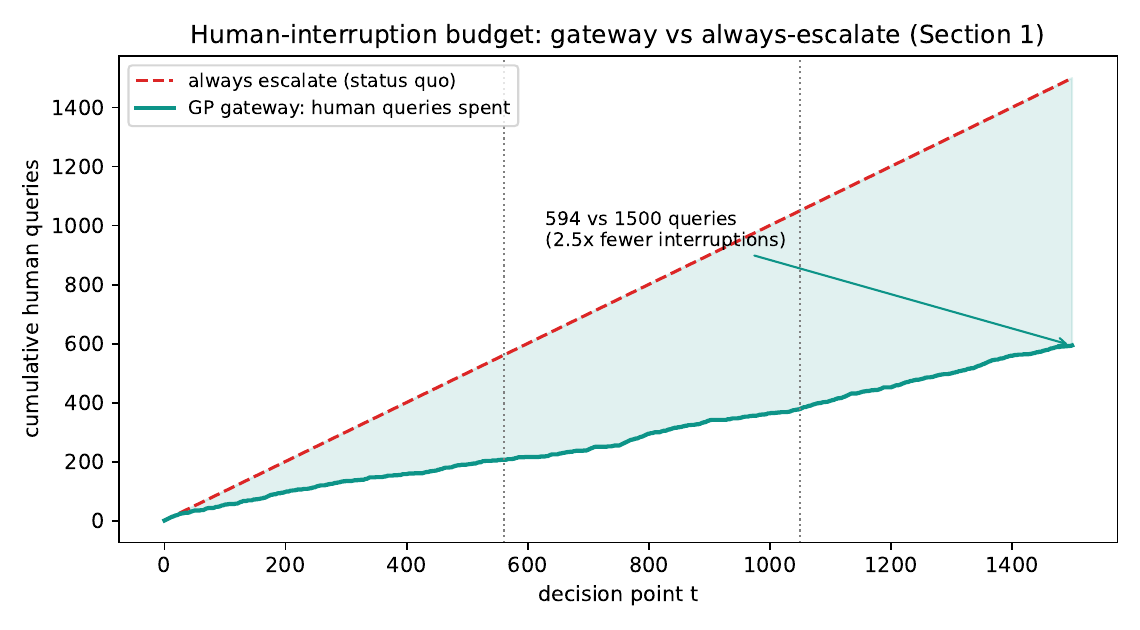}\hfill
\includegraphics[width=0.49\linewidth]{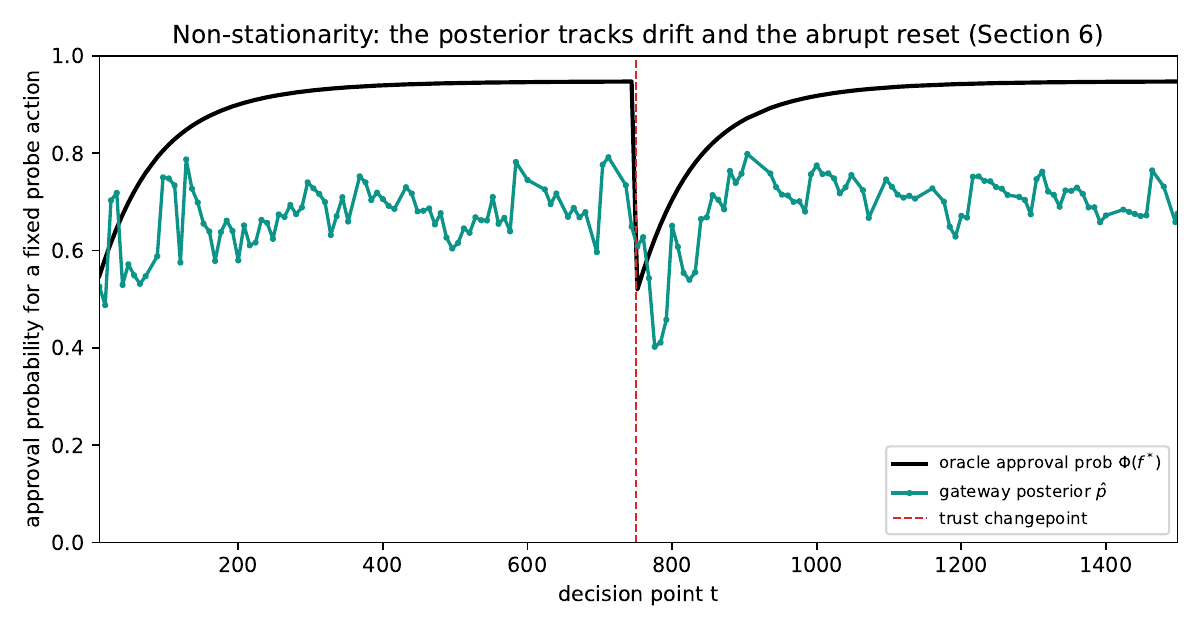}
\caption{Left: cumulative human queries, gateway versus always-escalate
(Section~1). Right: a fixed probe action's posterior approval probability
tracks the drifting oracle and its abrupt reset (Section~6); the systematic
gap is the Laplace-probit underconfidence noted in the text.}
\label{fig:drift}
\end{figure}

\paragraph{Inference and correlated generalization hold.}
Table~\ref{tab:results} reports the validation phase (a fair measurement of
what was learned, with the changepoint inside it) and the post-changepoint
test phase, against a no-correlation baseline that learns each tool
independently. The GP gateway auto-decides $68\%$ of actions at $97.3\%$
accuracy with a $2.4\%$ false-allow rate on validation, and $99.7\%$ accuracy
post-changepoint; the independent baseline auto-decides almost everything but
at $69$--$71\%$ accuracy, confirming that the kernel, not the three-tier rule,
carries the quality. Correlated generalization is decisive: on a held-out
\texttt{write\_file} to a test directory that was \emph{never} queried, the GP
recovers the correct decision $98.7\% \pm 2.9$ of the time purely by kernel
extrapolation from similar tools and targets, against $66.7\% \pm 47.1$ for
the independent learner (chance $50\%$); Figure~\ref{fig:gateway} (right)
visualizes the gap. The \textsc{ask} band narrows and the auto-approve rate
rises as the posterior concentrates, then both react sharply at the
changepoint and recover (Figure~\ref{fig:gateway}, left), and a fixed probe
action's posterior tracks the drift and the abrupt reset
(Figure~\ref{fig:drift}). Calibration is directionally correct but
underconfident at the kernel-far tail (validation ECE $0.19$ against the true
$\Phi(f^*)$), which we report rather than tune away.

\begin{table}[tb]
\centering
\small
\begin{tabular}{@{}lcccc@{}}
\toprule
& \multicolumn{2}{c}{\textbf{Validation (headline)}}
& \multicolumn{2}{c}{\textbf{Test (post-changepoint)}} \\
\cmidrule(lr){2-3}\cmidrule(lr){4-5}
\textbf{Metric} & GP gateway & Independent & GP gateway & Independent \\
\midrule
Auto-decision accuracy      & {\boldmath$0.973 \pm .016$} & $0.688 \pm .164$ & {\boldmath$0.997 \pm .003$} & $0.705 \pm .248$ \\
False-allow rate (safety)   & {\boldmath$0.024 \pm .018$} & $0.027 \pm .017$ & {\boldmath$0.003 \pm .003$} & $0.007 \pm .008$ \\
Auto-decided fraction       & $0.684 \pm .020$ & $1.000 \pm .000$ & $0.644 \pm .094$ & $0.933 \pm .099$ \\
\textsc{ask} / escalation   & $0.316 \pm .020$ & $0.000 \pm .000$ & $0.356 \pm .094$ & $0.067 \pm .099$ \\
prob-RMSE vs $\Phi(f^*)$    & {\boldmath$0.263 \pm .004$} & $0.344 \pm .076$ & {\boldmath$0.236 \pm .025$} & $0.353 \pm .102$ \\
ECE vs true prob            & {\boldmath$0.192 \pm .008$} & $0.273 \pm .078$ & {\boldmath$0.205 \pm .022$} & $0.316 \pm .104$ \\
\bottomrule
\end{tabular}
\caption{Simulation results, mean $\pm$ std over $6$ seeds. \textbf{Bold}
marks the decisive cells: the GP gateway is far more accurate, safer, and
better calibrated than a no-correlation contextual baseline. The baseline's
larger automated fraction is not a virtue (left unbolded): it automates more
only by auto-deciding cases it gets wrong.}
\label{tab:results}
\end{table}

\paragraph{Human-burden reduction.}
Against the status quo of always escalating (one human query per action), the
gateway spends $508 \pm 51$ queries over the scored phases versus $940$, a
$\sim\!1.8\times$ reduction in interruptions, while auto-deciding most actions
accurately and safely (Figure~\ref{fig:drift}, left; the full-stream
cumulative trajectory shows a larger gap because the cold-start learn phase
escalates heavily by design). This is the Section~1 promise, and it is the
defensible headline rather than a comparison to random querying.

\paragraph{An honest negative result on the acquisition rule.}
The rule ``query inside the \textsc{ask} band,'' taken literally as a
sample-efficiency claim, is \emph{not} supported here: at a matched query
budget its prequential boundary-decision accuracy is $76.5\% \pm 2.2$ against
$78.4\% \pm 4.2$ for random querying. An ablation that switches the changepoint
off and on (Table~\ref{tab:ablation}) shows the gap is non-positive in
\emph{both} regimes, including with a fully stationary target, so the deficit
is not caused by non-stationarity. It is the familiar behaviour of pure
uncertainty sampling under class imbalance: a region the posterior has grown
confident about leaves the \textsc{ask} band and is never re-probed, so its
estimate is never refreshed and a silently reset tolerance there goes
undetected. The time-decayed kernel down-weights stale evidence but does not
itself generate new probes. (The per-condition magnitude is small and
seed-noisy; the robust finding is the consistently non-positive sign.) A
recency-aware or information-theoretic acquisition criterion (an
expected-information or BALD rule with a forgetting term) is the natural
remedy and is left to future work; it does not affect the inference,
generalization, or burden-reduction results above, which use the same band as
a pure escalation rule.

\begin{table}[tb]
\centering
\small
\begin{tabular}{@{}lccc@{}}
\toprule
\textbf{Oracle} & \textbf{\textsc{ask}-band active} & \textbf{Random query}
& \textbf{Gap} \\
\midrule
Stationary (no changepoint)   & $80.6\%$ & $83.9\%$ & {\boldmath$-3.3$} pp \\
With \S\ref{sec:nonstationary} changepoint & $75.8\%$ & $76.2\%$ & {\boldmath$-0.4$} pp \\
\bottomrule
\end{tabular}
\caption{Acquisition ablation ($5$ seeds, matched budget, prequential
boundary accuracy). The \textbf{bold} gap is the finding: it is non-positive
in both regimes, so the \textsc{ask}-band rule does not beat random querying
even when the target is stationary; the deficit is generic, not a
changepoint artifact.}
\label{tab:ablation}
\end{table}

\section{Concluding Remarks}

We have argued that deciding when an agent may act autonomously is not a
threshold to hand-tune but a latent human risk-tolerance function to learn,
and that doing so is structurally an instance of Preferential Bayesian
Optimization specialized to unary approve/deny feedback. The contribution is
the mechanism the governance literature leaves implicit: a GP-probit policy
gateway whose allow/escalate/block boundary adapts from feedback, generalizes
across correlated actions through a structured kernel, and tracks a drifting
supervisor through a time-decayed component.

The simulation study supports the parts of this story that concern
\emph{inference}: the gateway recovers a non-stationary boundary it never
observes directly, transfers evidence to unqueried action-context
combinations far better than an independent learner, tracks an abrupt
changepoint, and cuts human interruptions substantially relative to escalating
everything. It also disciplines one claim: the \textsc{ask} band is a sound
\emph{escalation} rule, but treating it as a sample-efficient
\emph{acquisition} rule does not hold under class imbalance, independently of
non-stationarity. We report this rather than tune it away, and read it as
locating the open problem precisely in the acquisition criterion rather than
in the formulation. The remaining gap is empirical: no public dataset tracks a
single supervisor's per-action decisions as their tolerance drifts, so a
recency-aware acquisition rule and a longitudinal human study are the natural
next steps. We see the value of the formulation less in the present numbers,
which are simulated by necessity, than in turning a hand-set autonomy tier
into an object that can be learned, audited, and questioned.

\bibliographystyle{plainnat}
\bibliography{references}

\end{document}